\documentclass[10pt,twocolumn,letterpaper]{article}

\usepackage{iccv}              

\usepackage{colortbl}
\usepackage{xcolor} 
\definecolor{lightgray}{gray}{0.9} 
\usepackage{graphicx}
\usepackage{wrapfig}
\usepackage{booktabs} 

\definecolor{iccvblue}{rgb}{0.21,0.49,0.74}
\usepackage[pagebackref,breaklinks,colorlinks,allcolors=iccvblue]{hyperref}
\usepackage{adjustbox}
\usepackage{multirow}
\usepackage{algorithm}
\usepackage{algorithmic}

\def\paperID{12836} 
\def\confName{ICCV}
\def\confYear{2025}

\title{SAMO: A Lightweight Sharpness-Aware Approach for Multi-Task Optimization with Joint Global-Local Perturbation }

\author{Hao Ban, Gokul Ram Subramani, Kaiyi Ji\thanks{Corresponding author.}\\
University
at Buffalo\\
{\tt\small haoban@buffalo.edu, gsubrama@buffalo.edu, kaiyiji@buffalo.edu}
}

\begin{document}
\maketitle
\begin{abstract}
Multi-task learning (MTL) enables a joint model to capture commonalities across multiple tasks, reducing computation costs and improving data efficiency. However, a major challenge in MTL optimization is task conflicts, where the task gradients differ in direction or magnitude, limiting model performance compared to single-task counterparts. Sharpness-aware minimization (SAM) minimizes task loss while simultaneously reducing the sharpness of the loss landscape. Our empirical observations show that SAM effectively mitigates task conflicts in MTL. Motivated by these findings, we explore integrating SAM into MTL but face two key challenges. While both the average loss gradient and individual task gradients-referred to as global and local information-contribute to SAM, how to combine them remains unclear. Moreover, directly computing each task gradient introduces significant computational and memory overheads. To address these challenges, we propose SAMO, a lightweight \textbf{S}harpness-\textbf{A}ware \textbf{M}ulti-task \textbf{O}ptimization approach, that leverages a joint global-local perturbation. The local perturbations are approximated using only forward passes and are layerwise normalized to improve efficiency. Extensive experiments on a suite of multi-task benchmarks demonstrate both the effectiveness and efficiency of our method. Code is available at \url{https://github.com/OptMN-Lab/SAMO}.
\end{abstract}
\section{Introduction}

Multi-task learning (MTL) is a machine learning paradigm that aims to learn multiple tasks simultaneously \cite{crawshaw2020multi,zhang2021survey,vandenhende2021multi}. Instead of training each task independently, MTL encourages the model to capture shared patterns across tasks, thus improving data efficiency and enhancing generalization across all tasks \cite{ruder2017overview}. This approach is widely used in various domains, including natural language processing \cite{chen2024multi,yu2024unleashing,zhang2023survey}, computer vision \cite{dai2016instance,ciaparrone2020deep}, and speech recognition \cite{tseng2024av,wang2024viola}, among others.

One major challenge in the optimization of MTL is task conflict, where the gradients of different tasks have varying magnitudes or divergent directions \cite{chen2025gradient}. The conventional optimization objective in MTL is to minimize the average loss. However, this approach causes the task with the largest gradient magnitude to dominate the update, potentially overshadowing other tasks and hindering the overall performance. To mitigate this issue, various methods have been developed, primarily focusing on gradient manipulation \cite{kendall2018multi,sener2018multi,liu2021conflict,navon2022multi,liu2023famo,senushkin2023independent,xiao2023direction,ban2024fair}. These methods aim to determine a compromise update direction that reduces each task's loss while promoting a more balanced solution.


Sharpness-Aware Minimization (SAM)~\cite{foretsharpness} aims to update model parameters in a way that not only minimizes the task loss but also reduces the sharpness of the loss landscape. This approach is widely used to improve generalization and has been applied in various domains, including meta-learning~\cite{abbas2022sharp}, domain generalization~\cite{wang2023sharpness,cha2021swad}, and transfer learning~\cite{bahri2022sharpness}. However, its application to multi-task learning (MTL) remains largely unexplored except for the recent study by~\cite{phan2022improving}. 
 \citealt{phan2022improving} integrated SAM into MTL and proposed a method called F-MTL, which applies SAM individually to each task. Specifically, it decomposes each task’s gradient into a low-loss component and a flat-seeking component, then applies a separate gradient manipulation strategy to each. 
While F-MTL improves performance, it faces two main challenges. {\bf First}, the computational cost increases significantly, as applying SAM to individual tasks introduces $K$ additional gradient computations, and the separate manipulation of the two gradient components doubles both memory and time cost, where $K$ is the number of tasks. {\bf Second}, the perturbations in F-MTL exploit only {\bf task-specific} information, neglecting the {\bf shared} positive transfer across tasks.

\begin{figure}[t]
\vspace{-0.4cm}
  \centering
   \includegraphics[width=0.9\linewidth]{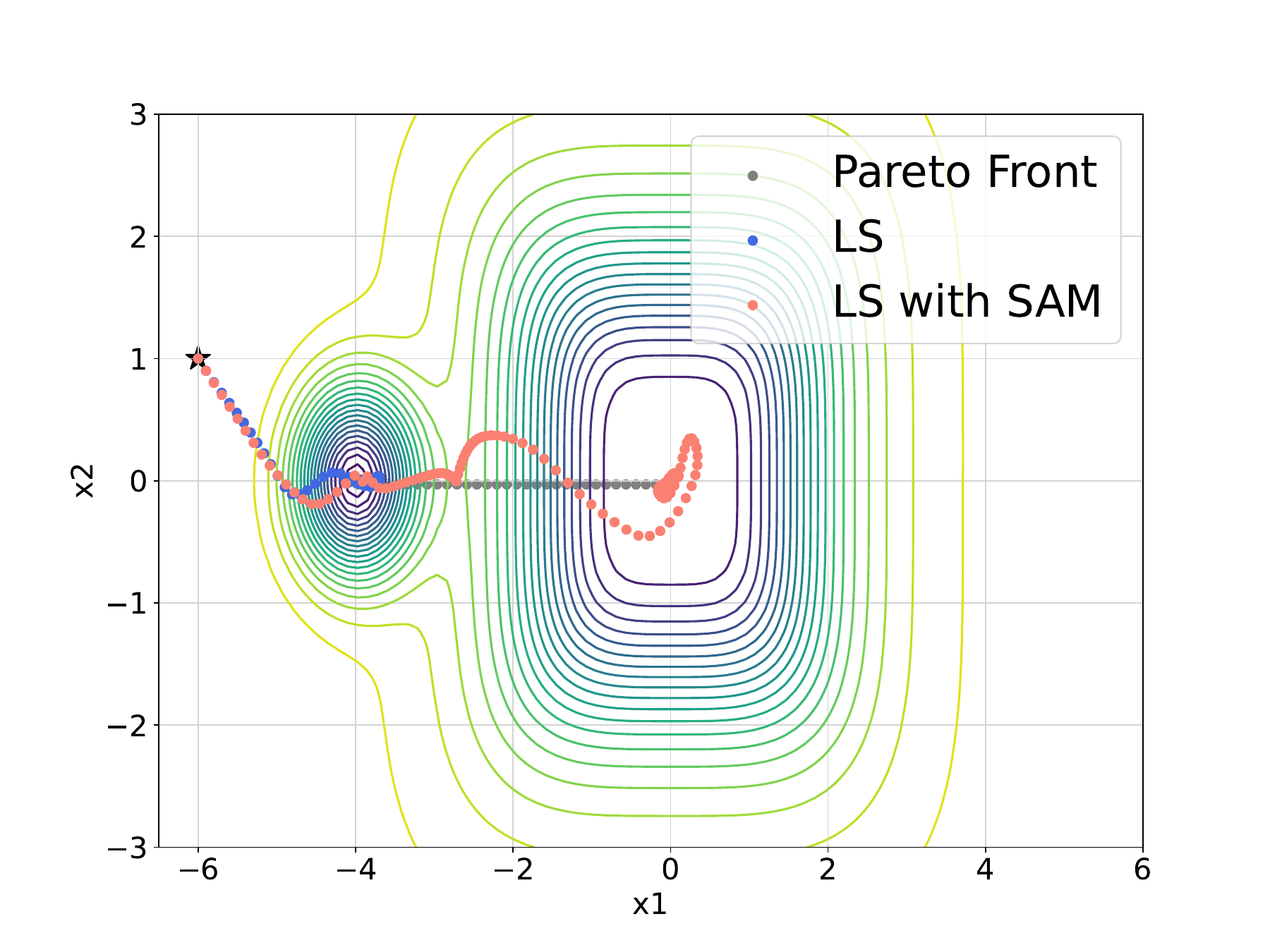}
      \vspace{-0.2cm}
   \caption{Optimization trajectories for a two-objective synthetic example using LS and LS with SAM, where the latter applies the gradient of the averaged loss as a perturbation. The $\star$ denotes the starting point. Both algorithms converge to the Pareto front. Furthermore, the optimization converges to a sharp minimum for LS, whereas LS with SAM leads to convergence at a flat minimum. {\em In this flat region, changes in one objective do not significantly affect the other.} SAM helps guide the model toward a broader convergence region, suggesting its effectiveness in mitigating task conflicts. Please refer to \Cref{exp:toy example} for full details. 
   }
   \label{fig:toy_example}
   \vspace{-0.3cm}
\end{figure}

In this paper, we tackle these challenges by introducing a lightweight, sharpness-aware approach for multi-task optimization that exploits both the task-specific and shared information. Our key contributions are summarized below.
\begin{list}{$\bullet$}{\topsep=0.3ex \leftmargin=0.25in \rightmargin=0.in \itemsep =-0.022in}
    \item We begin with a comprehensive  analysis of how SAM impacts MTL performance, using both a synthetic problem (see \Cref{fig:toy_example}) and real-world datasets (see \Cref{fig:cosine} and \Cref{tab:sharpness}). Our findings show that SAM encourages the model to converge to a broader region with reduced conflict while increasing the cosine similarity of task gradients, effectively mitigating task conflicts. Furthermore, we demonstrate that the average loss gradient and individual task gradients—referred to as global and local information, respectively—both play a role in conflict alleviation and performance enhancement. This motivates us to incorporate both types of information when integrating SAM into MTL.
    \item We propose lightweight {\bf S}harpness-{\bf A}ware {\bf M}ulti-task {\bf O}ptimization (\textbf{SAMO}). Our approach has two key features. {\bf First}, SAMO integrates both global and local information by computing a weighted average of task-specific and shared gradients as perturbations. {\bf Second}, instead of directly computing task-specific gradients, which incurs significant computational overheads, SAMO approximates them using a zeroth-order-like gradient estimation using only {\bf forward passes}. To further stabilize this zeroth-order computation, we introduce a novel layerwise normalization strategy. Differently from F-MTL that requires extra $K$ backpropagations, SAMO introduces only {\bf forward passes}, making substantially reduced time and memory. 
    
    \item SAMO is flexible and easily integrated into existing stat-of-the-art  gradient manipulation methods. We conduct extensive experiments on diverse real-world datasets including Cityscapes, NYUv2, CelebA, QM9, and Office Home, covering both single-input and multi-input settings.  
   Our empirical results demonstrate that SAMO consistently enhances performance across various multi-task optimization methods. For example, when integrated with FairGrad, SAMO improves the state-of-the-art results from 3.9 to -0.62 on Cityscapes, from -4.96 to -6.55 on NYU-v2, and from 0.37 to -0.74 on CelebA. Furthermore, it significantly outperforms methods that utilize only global or local perturbation information, underscoring the effectiveness of our proposed joint global-local perturbations. Finally, our runtime comparison reveals that SAMO is 
   substantially faster than the existing multi-task SAM approaches such as F-MTL.
\end{list}

\section{Related Work}

{\bf Multi-task learning.} 
The primary challenge in MTL lies in managing task interactions—promoting positive knowledge transfer while mitigating negative conflicts.

One class of approaches focus on novel architecture designs. Soft-parameter-sharing methods regularize parameters to be similar but not identical, allowing flexibility for related but distinct tasks \cite{gao2019nddr,ruder2019latent}. Modularity-based methods leverage mixture-of-experts (MoE) architectures to mitigate task conflicts by selectively activating task-specific experts \cite{chen2023mod,fan2022m3vit}, or employ neural architecture search to determine optimal branching points, enabling task-specific module learning \cite{gao2020mtl,bruggemann2020automated}.

Another line of research explores MTL optimization techniques to address task conflicts through task balancing. These methods include heuristic task loss weighting \cite{linreasonable,kendall2018multi,xiao2025scalable}, gradient balancing, which eliminates conflicting gradient components \cite{chen2018gradnorm,yu2020gradient,javaloyrotograd,senushkin2023independent}, and gradient weighting—the focus of this work. \citealt{desideri2012multiple} introduced MGDA, a multi-objective optimization method ensuring convergence to the Pareto front, which \citealt{sener2018multi} later adapted for MTL in deep neural networks. Since then, various MGDA extensions have been proposed to balance gradient conflicts and overall performance \cite{liu2021conflict,fernandomitigating,xiao2023direction,navon2022multi,ban2024fair,wang2025theoretical,zhang2024convergence}.

\noindent{\bf Sharpness-aware minimization.}
The relationship between the geometry of the loss landscape and generalization has been extensively studied \cite{hochreiter1994simplifying,hochreiter1997flat,dinh2017sharp,keskar2017large,li2018visualizing,jiangfantastic}, forming the foundation for methods that target flat minima. \citealt{foretsharpness} introduced SAM as a direct approach to optimizing sharpness, achieving notable generalization improvements across various tasks. Subsequent research has refined SAM by incorporating surrogate gaps \cite{zhuangsurrogate} and enhancing training efficiency \cite{duefficient,liu2022towards}, among other advancements.

The success of SAM has also inspired theoretical investigations into its generalization properties. \citealt{andriushchenko2022towards} analyzed its implicit bias in diagonal linear networks, while \citealt{wen2023sharpness} demonstrated that SAM reduces the largest eigenvalue of the Hessian in full-batch training. \citealt{chen2023does} compared the training dynamics of two-layer convolutional ReLU networks under SGD and SAM, showing that SAM can lead to benign overfitting in cases where SGD suffers from harmful overfitting.

Beyond theoretical insights, several studies have explored the empirical effects of SAM. \citealt{andriushchenko2023sharpness} found that SAM encourages low-rank representations in deep networks, while \citealt{springersharpness} showed that SAM implicitly promotes balanced feature learning, thereby improving feature quality. Our work builds on these findings from an MTL perspective, demonstrating that SAM effectively mitigates task conflicts. Furthermore, we introduce a novel sharpness-aware MTL approach that incorporates both global and local perturbations.

\section{Preliminaries}
In this section, we provide a brief overview of the optimization of MTL, then we introduce SAM.

\subsection{Multi-task optimization}
MTL aims to optimize multiple objective functions as 
\begin{align*}
    \min_{\theta\in\mathbb{R}^m} L=(l_1(\theta),l_2(\theta),\cdots,l_K(\theta)),
\end{align*} 
where $K$ denotes the number of tasks, and $\{l_i\}_{i=1}^K$ are the objectives parameterized by $\theta \in \mathbb{R}^m$. Given two parameter points $\theta_1$ and $\theta_2$, if $L(\theta_1) \neq L(\theta_2)$ and for all $i \in [K]$, we have $l_i(\theta_1) \leq l_i(\theta_2)$, then $\theta_1$ is said to dominate $\theta_2$.

A point is called Pareto optimal if it is not dominated by any other point—that is, no other solution simultaneously achieves better performance across all objectives. A point is termed Pareto stationary if there exists a convex combination of gradients of all objectives that sums to zero, indicating linear dependence. Pareto stationarity is a necessary condition for Pareto optimality, and most existing optimization methods in MTL aim to find Pareto stationary points.

\subsection{Sharpness-aware minimization}
Given a model parameterized by $\theta \in \mathbb{R}^m$ and a loss function $l(\theta)$, consider a small perturbation $\epsilon$ applied to the model parameters, where $\|\epsilon\| \leq \rho$. The resulting change in loss, $l(\theta + \epsilon) - l(\theta)$, serves as an indicator of the sharpness of the loss landscape at $\theta$ in the direction of $\epsilon$. The SAM algorithm is designed to simultaneously minimize both the loss and the sharpness of the loss landscape, promoting better generalization \cite{foretsharpness}.
\begin{align*}
    \min_{\theta\in\mathbb{R}^m} \max_{\|\epsilon\|\le\rho} l(\theta+\epsilon)=[l(\theta+\epsilon)-l(\theta)]+l(\theta),
\end{align*}
 where the inner maximization problem, $\max_{\|\epsilon\|\le\rho}l(\theta+\epsilon)$, can be approximated using its first-order Taylor expansion. This leads to the perturbation estimate as 
 \begin{align*}
     \hat{\epsilon}(\theta)=\rho\nabla{l(\theta)}/\|\nabla{l(\theta)}\|,
 \end{align*}
which indicates that the perturbation moves in the direction of the current gradient with a small step forward. The gradient of the outer minimization problem is then given by 
\begin{align*}
\nabla_\theta l(\theta+\hat{\epsilon}(\theta))&=\frac{d(\theta+\hat{\epsilon}(\theta))}{d\theta}\cdot\nabla_\theta l(\theta)\big\vert_{\theta+\hat{\epsilon}(\theta)} \\
&\approx \nabla_\theta l(\theta)\big\vert_{\theta+\hat{\epsilon}(\theta)},
\end{align*}
where the approximation drops second-order gradients to reduce the computational cost. 


\section{SAM Mitigates Task Conflicts}\label{sec:sam_conflict}



In this section, we systematically analyze the impact of SAM on MTL from three perspectives: (i) mitigation of task conflicts, (ii) different types of sharpness information, and (iii) challenges in applying SAM to MTL.

\subsection{Sharpness and task conflicts}
We first investigate if SAM is effective in mitigating conflicts among different tasks. We 
analyze the cosine similarities among task gradients before and after applying SAM. We compute the Hessian spectrum, where the maximum eigenvalue $\lambda_{\text{max}}$ and the bulk of the spectrum, measured as the ratio of the maximum and the fifth largest eigenvalue $\lambda_{\text{max}}/\lambda_5$, can be used as proxy for sharpness \cite{foretsharpness,jastrzebskibreak,yao2018hessian}. 

\begin{figure}
    \centering
    \includegraphics[width=0.535\linewidth]{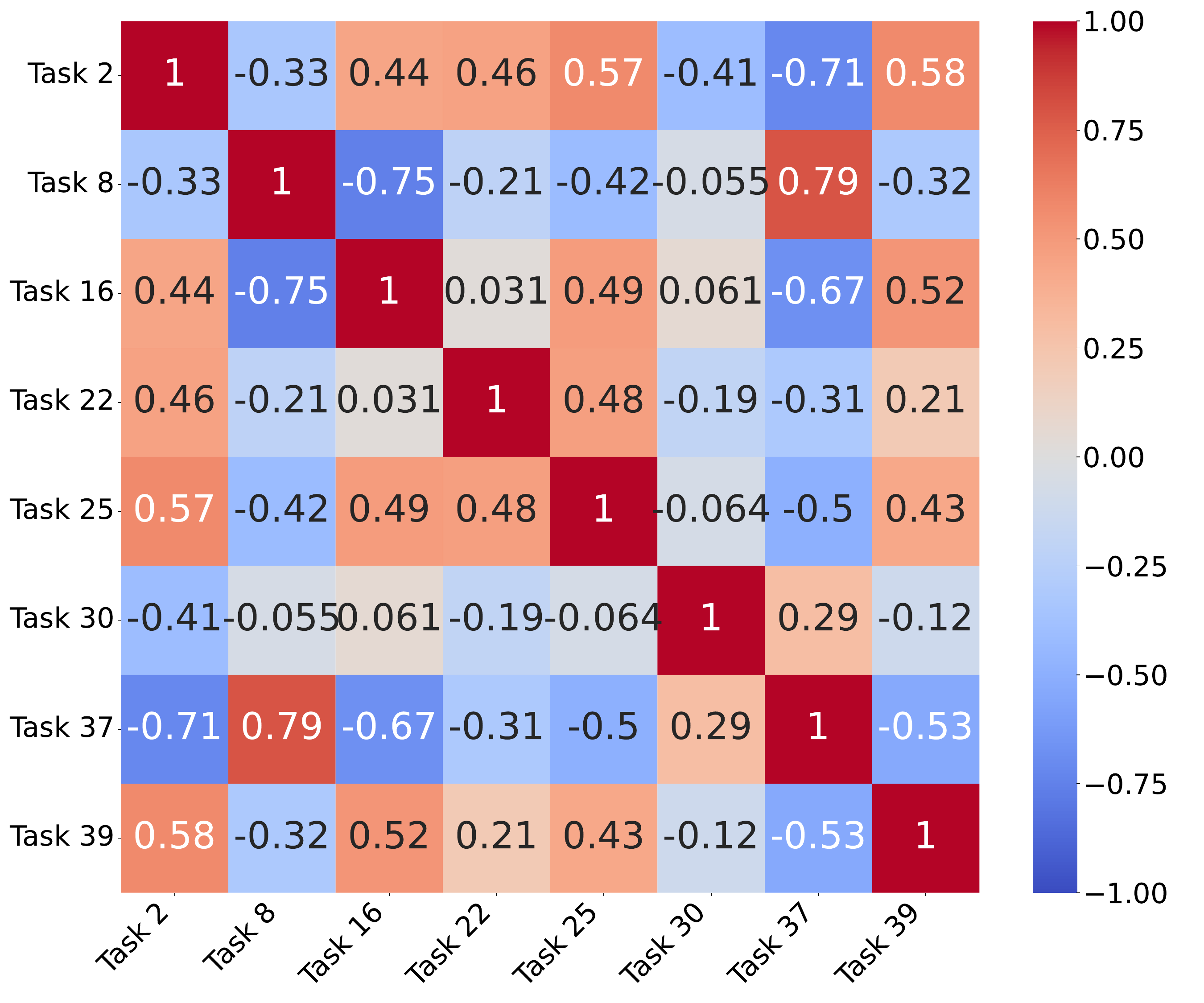}
    \hspace{-8.5mm}
    \includegraphics[width=0.535\linewidth]{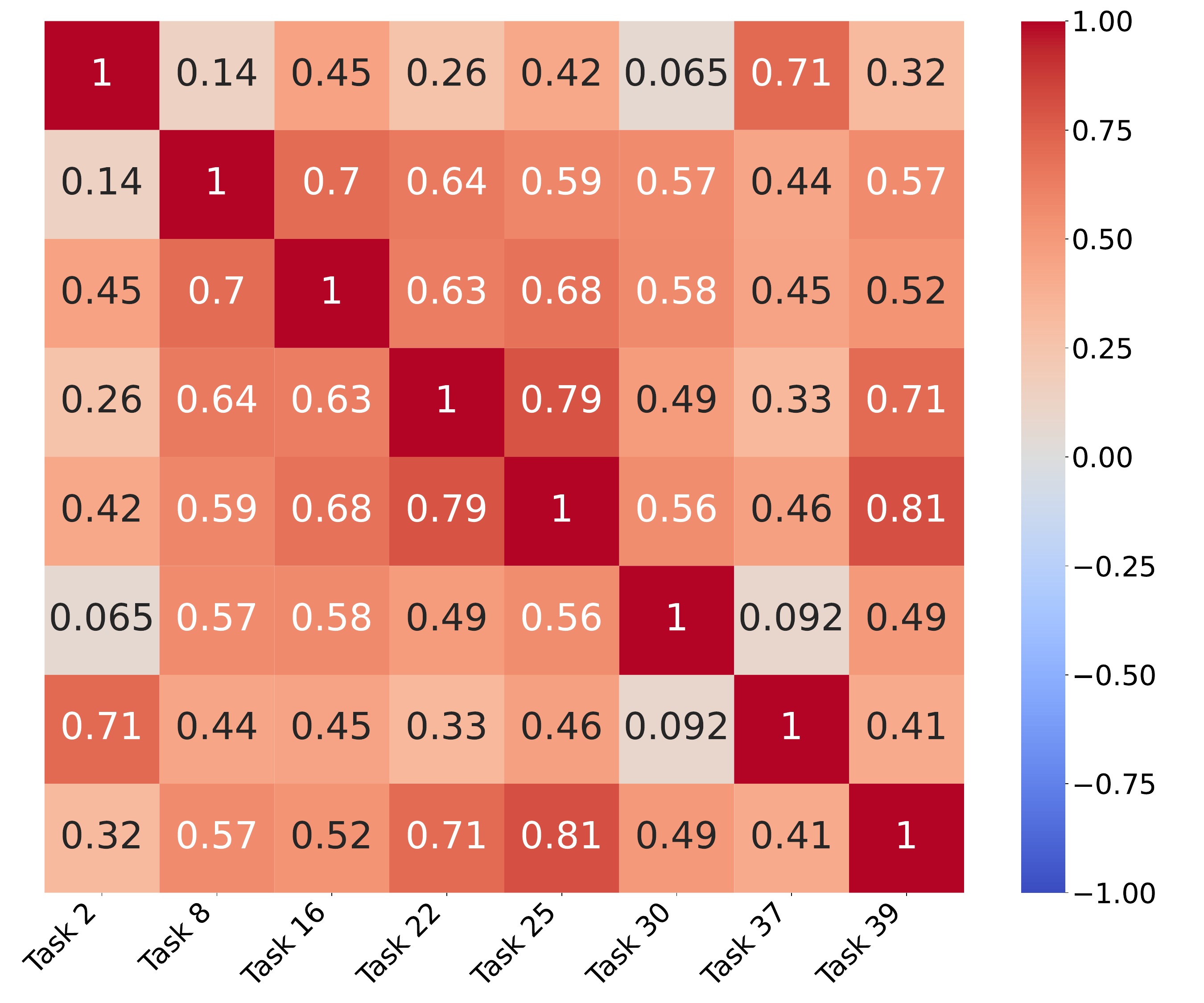}
    \caption{Cosine similarities of task gradients for LS (left) and LS with SAM (right). The latter achieves higher cosine similarities, indicating that task conflicts have been significantly mitigated.}
    \label{fig:cosine}
\end{figure}

\begin{table}[t]
  \centering
  \begin{adjustbox}{width=\linewidth}
  \begin{tabular}{ccccccc}
    \toprule
    \multirow{2}*{Epoch} & \multicolumn{3}{c}{LS} & \multicolumn{3}{c}{LS with SAM}\\
    \cmidrule(lr){2-4}\cmidrule(lr){5-7}
    & $\lambda_{\text{max}}$ & $\lambda_{\text{max}}/\lambda_5$ & Avg Acc &  $\lambda_{\text{max}}$ & $\lambda_{\text{max}}/\lambda_5$ & Avg Acc \\
    \midrule
    5 & 235.9 & 16.9 & 67.80 & 128.2 & 8.5 & 67.63\\
    10 & 493.5 & 32.9 & 74.96 & 155.0 & 10.3 & 76.44 \\
    15 & 471.7 & 31.4 & 76.17 & 168.1 & 11.2 & 77.15 \\
    \bottomrule
  \end{tabular}
  \end{adjustbox}
  \caption{The sharpness of loss landscape and model performance. Reduced sharpness leads to improved performance.}
  \label{tab:sharpness}
  \vspace{-0.15cm}
\end{table}

We conduct experiments on the CelebA dataset, randomly selecting 8 out of 40 tasks. We compare the LS with global SAM, which computes the perturbation term using the gradient of the averaged losses. \Cref{fig:cosine} illustrates the cosine similarities of task gradients after the 15th epoch, while \Cref{tab:sharpness} reports the sharpness of the loss landscape along with the corresponding performance. Notably, LS with SAM achieves a significantly flatter loss landscape and a higher accuracy compared to LS. Moreover, the reduction in sharpness leads to higher cosine similarity among task gradients and improved performance. These findings suggest that SAM mitigates task conflicts by flattening the loss landscape, thereby enhancing the effectiveness of MTL methods.

\subsection{Both global and local information help}\label{sec:glo-loc}

In this section, we analyze the impact of two types of sharpness information on MTL performance. The first approach, which incorporates {\bf global information}, computes the gradient based on the average losses and applies a single shared perturbation term across all tasks. In contrast, the second approach, referred to as SAM with {\bf local information}, calculates the gradient separately for each task, resulting in task-specific perturbation terms.

The results are presented in \Cref{tab:global_local_sam}. Both global and local information significantly enhance the performance of vanilla MTL methods. However, it remains unclear which is more effective. For example, G-SAM outperforms L-SAM when combined with FairGrad on NYU-v2, whereas L-SAM surpasses G-SAM when paired with MGDA on Cityscapes. These findings suggest that integrating both types of information into MTL can lead to greater performance improvements.


\begin{table}
  \small
  \centering
  \begin{adjustbox}{width=0.85\linewidth}
  \begin{tabular}{lccc}
    \toprule
    \multirow{2}*{Method} & \multicolumn{1}{c}{Cityscapes} & \multicolumn{1}{c}{NYU-v2} & \multicolumn{1}{c}{CelebA} \\ 
     
    \cmidrule(lr){2-2}\cmidrule(lr){3-3}\cmidrule(lr){4-4}
    & $\Delta m\%\downarrow$ & $\Delta m\%\downarrow$ & $\Delta m\%\downarrow$ \\
    \midrule
    LS & 22.60 & 5.59 & 4.15 \\
    G-SAM-LS & 17.85 & 3.85 & 3.24  \\
    L-SAM-LS & \cellcolor{lightgray} 16.71 & \cellcolor{lightgray} 2.11 & \cellcolor{lightgray} 2.90  \\
    \midrule
    MGDA & 44.14 & 1.38 & 14.85 \\
    G-SAM-MGDA &\cellcolor{lightgray} 7.51 & \cellcolor{lightgray}-0.23 & 11.78  \\
    L-SAM-MGDA & 11.94 & 0.01 & \cellcolor{lightgray}8.47\\
    \midrule
    FairGrad & 3.90 & -4.96 & 0.37  \\
    G-SAM-FairGrad & \cellcolor{lightgray} 0.93 & \cellcolor{lightgray} -5.70 & 0.41 \\
    L-SAM-FairGrad & 1.01 & -5.42 &\cellcolor{lightgray} -0.42 \\
    \bottomrule
  \end{tabular}
  \end{adjustbox}
  \caption{Comparison of SAM with global information (G-SAM) and local information (L-SAM). $\uparrow$ indicates that higher values are better, while $\downarrow$ denotes the opposite. The results show that both global and local information enhances SAM's effectiveness.}
  \label{tab:global_local_sam}
  \vspace{-0.5cm}
\end{table}



\subsection{Challenges to use joint global-local information}

Although both global and local information are important, how to effectively combine and integrate them into SAM and MTL frameworks remains unclear. To the best of our knowledge, no prior work has addressed this problem in the MTL context.

Another challenge is the additional computational cost of leveraging local information. Specifically, local SAM computes perturbations separately for each task, resulting in $K$ {\bf additional} gradient computations and backpropagations, where $K$ is the number of tasks. 
Thus, developing a lightweight SAM method that avoids this additional $\mathcal{O}(K)$ overhead is essential.

\section{Our Proposed Method: SAMO}

\begin{figure*}[t]
  \centering
   \includegraphics[width=0.9\linewidth]{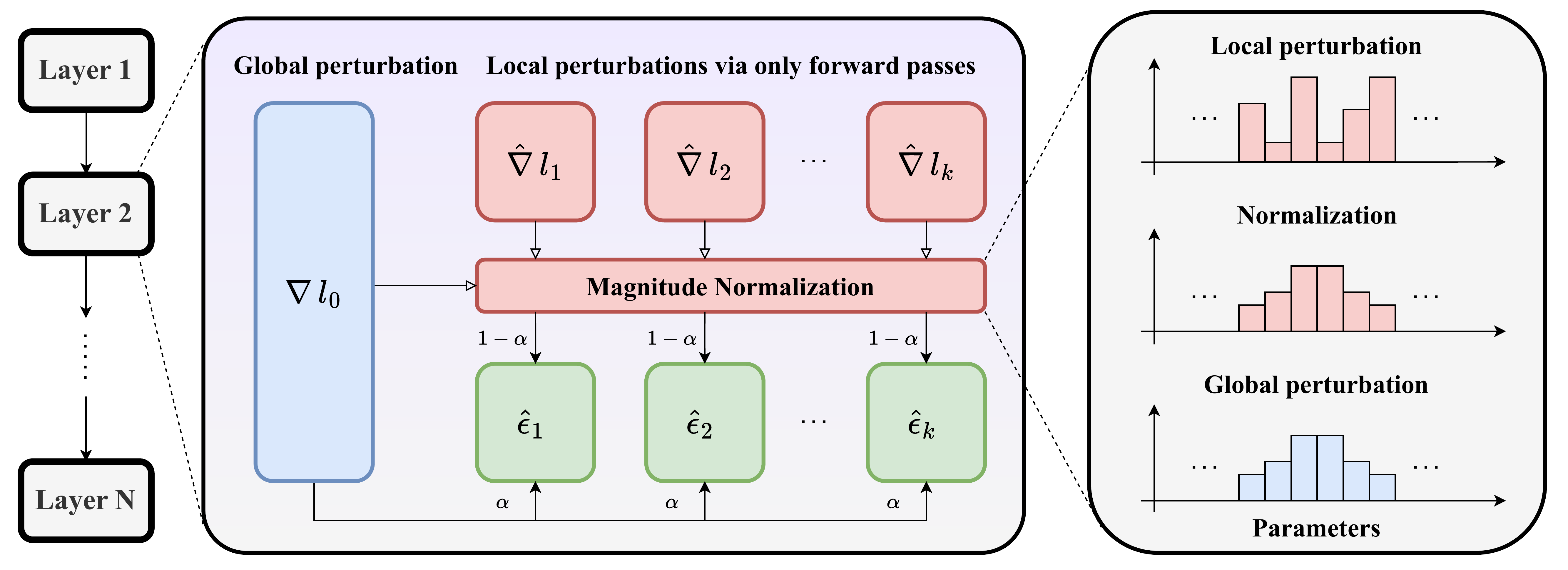}\
   \hspace{20mm}

      \vspace{-0.7cm}

   \caption{Our SAMO computes joint global and local perturbations for all tasks. Left: The joint perturbation is obtained by a weighted average of global and local information. Right: Layerwisely magnitude normalization. Local perturbations are layerwise normalized to match the magnitude of global perturbation.
   } 
   \label{fig:gradient_normalization}
      \vspace{-0.2cm}
\end{figure*}

Inspired by the empirical findings in \Cref{sec:sam_conflict}, we propose a lightweight \textbf{S}harpness-\textbf{A}ware \textbf{M}ulti-task \textbf{O}ptimization (SAMO) approach, which combines the benefits of G-SAM and L-SAM to mitigate task conflicts while keeping the computation cost manageable. 
As presented in \cref{alg:sharpmtl} and \Cref{fig:gradient_normalization}, SAMO consists of two key modules: (i) a joint global and local perturbation computation; (ii) an efficient zeroth-order local gradient approximation with layerwise normalization. 


\subsection{Joint global and local perturbations}
We first compute the global gradient $\nabla_\theta l_0(\theta)$ of the average loss $l_0=\frac{1}{K}\sum_{i=1}^K l_i$, along with the individual task gradients $\nabla_\theta l_i(\theta),i=1,...,K$. Then, the first stage of SAMO introduces the following perturbation for each task:
\begin{align}\label{eq:epsilon}
    \hat{\epsilon}_i(\theta)=\rho\frac{\alpha \nabla_\theta l_0(\theta)+(1-\alpha)\nabla_\theta l_i(\theta)}{\|\alpha \nabla_\theta l_0(\theta)+(1-\alpha)\nabla_\theta l_i(\theta)\|},
\end{align}
where $\alpha\in[0,1]$ is a tunable weighting scalar that balances the trade-off between global and local perturbations. In the second stage, the SAMO gradient is obtained for each task:
\begin{align}\label{eq:sam_gradient}
    g_i^{SAMO}=\nabla_\theta l_i(\theta+\hat{\epsilon}_i(\theta))\approx \nabla_\theta l_i(\theta)\big\vert_{\theta+\hat{\epsilon}_i(\theta)}.
\end{align}
This joint perturbation leverages the positive transfer induced by the globally averaged gradient while simultaneously capturing task-specific variations through local individual gradients.
\begin{algorithm}[t]
\small
   \caption{SAMO with joint global-local perturbations}
   \label{alg:sharpmtl}
\begin{algorithmic}[1]
   \STATE {\bf Input:} Model parameters $\theta_0$, loss functions $l_1,\cdots,l_K$, gradient manipulation MTL method $\mathcal{M}$, learning rate $\eta$, perturbation step size $\rho$, iteration steps $T$.
   \STATE {\bf Output:} MTL model trained with SAMO
   \FOR{$t=0$ {\bf to} $T-1$}
   \STATE Compute average gradient $\nabla_\theta l_0(\theta_t)$
     \FOR{task $i=1$ {\bf to} $K$}
     \STATE Compute layerwise gradient $\hat{\nabla}_\theta l_i(\theta_t)$ by \cref{eq:rescale}
     \STATE Compute $\hat{\epsilon}_i(\theta_t)$ by \cref{eq:epsilon}
     \STATE Compute gradient for SAMO $g_{t,i}^{SAMO}$ by \cref{eq:sam_gradient}
     \ENDFOR
   \STATE Compute $d_t=\mathcal{M}(g_{t,1}^{SAMO}, \cdots,g_{t,K}^{SAMO})$
   \STATE Update the parameters $\theta_{t+1}=\theta_t-\eta d_t$
   \ENDFOR
\end{algorithmic}
\end{algorithm}


 \subsection{Layerwise normalized local perturbation with only forward passes}


However, computing the \( K \) local gradients \( \{\nabla_\theta l_i(\theta)\}_{i=1}^K \) in \Cref{eq:epsilon} incurs substantial time and memory overhead. To mitigate this, we seek efficient and effective approximations for the local perturbations. Our approach is inspired by parameter-efficient fine-tuning (PEFT) techniques such as LoRA \cite{hulora}, where the pre-trained model remains shared while distinct LoRA modules adapt to different tasks. Specifically, we treat the global perturbation as a base and augment it with approximated local perturbations for each task, enabling efficient task-specific adaptations.

Our idea is to employ the stochastic perturbation simultaneous approximation (SPSA) gradient estimator, which relies solely on the forward pass computations \cite{spall1992multivariate}: 
\begin{align}\label{eq:zeroth_order}
    \hat{\nabla}_\theta l_i(\theta)\approx \frac{l_i(\theta+\mu z_i)-l_i(\theta - \mu z_i)}{2\mu}z_i,
\end{align}
where $\hat{\nabla}$ denotes the gradient estimator, $z_i\in\mathbb{R}^m$ is a random vector sample from a standard Gaussian distribution, and $\mu>0$ is a perturbation factor. Given the complexity of the loss landscape, the variance of the estimated gradient can fluctuate \cite{gautamvariance}, leading to instability during training. To address this, we propose normalizing the estimated gradient layerwise to match the magnitude of the average gradient, which is considered as a reference:
\begin{align}\label{eq:rescale}
    \hat{\nabla}_\theta l_i(\theta^d)\leftarrow\hat{\nabla}_\theta l_i(\theta^d)\frac{\|\nabla_\theta l_0(\theta^d)\|}{\|\hat{\nabla}_\theta l_i(\theta^d)\|},
\end{align}
where $\theta^d$ means the parameters of network layer at depth $d$. 

Thus, compared to general MTL methods, the total computation cost of our SAMO involves $K+1$
\begin{wraptable}{r}{.2\textwidth}
    \small
    \begin{minipage}[ht]{.2\textwidth}
        \centering
    \begin{tabular}{lc}
        \toprule
        Method  & Overheads \\
        \midrule
        G-SAM  & {\color{blue}$\mathbf{C_b}$} \\
        L-SAM  & $K{\bf \color{blue}C_b}$ \\
        F-MTL  & $K{\bf \color{blue}C_b} + C_{gm}$ \\
        SAMO   & ${\bf \color{blue}C_b} + 2K{\bf \color{gray}C_f}$ \\
        \bottomrule
    \end{tabular}
    \vspace{-0.25cm}
    \caption{Additional computational cost  for  $K$ tasks. Forward cost $C_f\ll$  Backward cost $C_b$. }
    \label{tab:overhead}
   \end{minipage}
    \vspace{-1.0em}
\end{wraptable}
 gradient computations along with additional forward pass computations, making it significantly more efficient than F-MTL, which directly applies SAM to each task \cite{phan2022improving}. Let $C_f$, $C_b$ and $C_{gm}$ represent the computation cost of a forward pass, a backward pass, and the gradient manipulation MTL method, respectively. The comparison of computational overheads is presented in \Cref{tab:overhead}. Note that the forward cost $C_f$ is marginal compared to $C_b,C_{gm}$, indicating that the overall complexity remains dominated by $C_b$. Thus, SAMO uses both global and local information while keeping costs on par with G-SAM that relies solely on global information.


\section{Experiments}\label{sec:experiments}
In this section, we first present a toy example to analyze how SAM mitigates task conflicts through the lens of optimization trajectory. We then evaluate the performance of the proposed SAMO on a range of real-world MTL datasets, covering both single-input and multi-input scenarios. In the single-input MTL, each input data sample has multiple labels, whereas in the multi-input MTL, each task has its own distinct dataset.

\subsection{Synthetic toy example}\label{exp:toy example}
We consider a 2-objective toy example with the following objective functions:
\begin{align*}
    f_1&=1-\frac{1}{1+\frac{1}{10}(x_1^4+x_2^4)} \\
    f_2&=1-\exp(-2(x_1+4)^2-2x_2^2),
\end{align*}
where the parameters are constrained to $x_1\in[-6,6]$, and $x_2\in[-3,3]$. In dynamic weighting MTL methods such as MGDA and FairGrad, task weights vary at each optimization step, altering the loss landscape. We use Linear Scalarization (LS), a static weighting method that directly sums the two objectives to facilitate a clearer analysis of the optimization trajectory in a static loss landscape. We then apply SAM with global information to LS, and compare the optimization trajectory of both methods. 

The results are shown in \Cref{fig:toy_example}. Both trajectories start from $(-6, 1)$, but LS and LS with SAM converge to different Pareto optimal points. While LS converges to a sharp region, LS with SAM reaches a flatter region. In these regions, as the parameters move, a decrease in one objective does not significantly increase the other, indicating a relatively low level of conflict between the two objectives. Notably, LS with SAM converges to a significantly larger flat region compared to LS, demonstrating that SAM effectively mitigates task conflicts.


\begin{table}
  \centering
  \begin{adjustbox}{width=\linewidth}
  \begin{tabular}{llllll}
    \toprule
    \multirow{2}*{Method} & \multicolumn{2}{c}{Segmentation} & \multicolumn{2}{c}{Depth} & 
    \multirow{2}*{$\Delta m\%\downarrow$} \\
    \cmidrule(lr){2-3}\cmidrule(lr){4-5}
    & mIoU $\uparrow$ & Pix Acc $\uparrow$ & Abs Err $\downarrow$ & Rel Err $\downarrow$ & \\
    \midrule
    STL & 74.01 & 93.16 & 0.0125 & 27.77 \\
    \midrule
    LS & 75.18 & 93.49 & 0.0155 & 46.77 & 22.60 \\
    SI & 70.95 & 91.73 & 0.0161 & 33.83 & 14.11 \\
    RLW & 74.57 & 93.41 & 0.0158 & 47.79 & 24.38 \\
    DWA & 75.24 & 93.52 & 0.0160 & 44.37 & 21.45 \\
    UW & 72.02 & 92.85 & 0.0140 & 30.13 & 5.89 \\
    MGDA & 68.84 & 91.54 & 0.0309 & 33.50 & 44.14 \\
    PCGrad & 75.13 & 93.48 & 0.0154 & 42.07 & 18.29 \\
    GradDrop & 75.27 & 93.53 & 0.0157 & 47.54 & 23.73 \\
    IMTL-G & 75.33 & 93.49 & 0.0135 & 38.41 & 11.10 \\
    CAGrad & 75.16 & 93.48 & 0.0141 & 37.60 & 11.64 \\
    MoCo & 75.42 & 93.55 & 0.0149 & 34.19 & 9.90 \\
    Nash-MTL & 75.41 & 93.66 & \cellcolor{lightgray}\textbf{0.0129} & 35.02 & 6.82 \\
    FAMO & 74.54 & 93.29 & 0.0145 & 32.59 & 8.13 \\
    FairGrad & 74.10 & 93.03 & 0.0135 & 29.92 & 3.90 \\
    F-MTL$^\sharp$ & 73.77 & 93.12 & \cellcolor{lightgray}\textbf{0.0129} & 27.44 & 0.67 \\
    \midrule
    SAMO-LS & \cellcolor{lightgray}\textbf{76.46} & \cellcolor{lightgray}\textbf{93.76} & 0.0147 & 39.85 & 14.30 \\
    SAMO-MGDA & 73.28 & 93.26 & 0.0133 & 30.57 & 4.30 \\
    SAMO-FairGrad & 74.37 & 93.14 & \cellcolor{lightgray}\textbf{0.0129} & \cellcolor{lightgray}\textbf{26.30} & \cellcolor{lightgray}\textbf{-0.62}$^*$ \\
    \bottomrule
  \end{tabular}
  \end{adjustbox}
  \caption{Results on Cityscapes (2-task) dataset. The best results are highlighted in \textbf{bold} with gray background. $^*$ indicates the best $\Delta m\%$ result. $^\sharp$ denotes the best results obtained by F-MTL.}
  \label{tab:cityscapes}
\end{table}

\subsection{Single-input MTL}\label{exp:single input}
We conduct experiments on the commonly used datasets described as follows. We then discuss the evaluation and the performance achieved by the proposed SAMO. Our implementation is based on the codebase released by Nash-MTL \cite{navon2022multi}.

\begin{table*}
  \centering
  \begin{adjustbox}{width=0.80\linewidth}
  \begin{tabular}{llllllllllc}
    \toprule
    \multirow{3}*{Method} & \multicolumn{2}{c}{Segmentation} & \multicolumn{2}{c}{Depth} & \multicolumn{5}{c}{Surface Normal} & \multirow{3}*{$\Delta m\%\downarrow$} \\
    \cmidrule(lr){2-3}\cmidrule(lr){4-5}\cmidrule(lr){6-10}
    & \multirow{2}*{mIoU $\uparrow$} & \multirow{2}*{Pix Acc $\uparrow$} & \multirow{2}*{Abs Err $\downarrow$} & \multirow{2}*{Rel Err $\downarrow$} & \multicolumn{2}{c}{Angle Distance $\downarrow$} & \multicolumn{3}{c}{Within $t^\circ$ $\uparrow$} & \\
    \cmidrule(lr){6-7}\cmidrule(lr){8-10}
    & & & & & Mean & Median & $<$11.25 & $<$22.5 & $<$30 & \\
    \midrule
    STL & 38.30 & 63.76 & 0.6754 & 0.2780 & 25.01 & 19.21 & 30.14 & 57.20 & 69.15 & \\
    \midrule
    LS & 39.29 & 65.33 & 0.5493 & 0.2263 & 28.15 & 23.96 & 22.09 & 47.50 & 61.08 & 5.59 \\
    SI & 38.45 & 64.27 & 0.5354 & 0.2201 & 27.60 & 23.37 & 22.53 & 48.57 & 62.32 & 4.39 \\
    RLW & 37.17 & 63.77 & 0.5759 & 0.2410 & 28.27 & 24.18 & 22.26 & 47.05 & 60.62 & 7.78 \\
    DWA & 39.11 & 65.31 & 0.5510 & 0.2285 & 27.61 & 23.18 & 24.17 & 50.18 & 62.39 & 3.57 \\
    UW & 36.87 & 63.17 & 0.5446 & 0.2260 & 27.04 & 22.61 & 23.54 & 49.05 & 63.65 & 4.05 \\
    MGDA & 30.47 & 59.90 & 0.6070 & 0.2555 & 24.88 & 19.45 & 29.18 & 56.88 & 69.36 & 1.38 \\
    PCGrad & 38.06 & 64.64 & 0.5550 & 0.2325 & 27.41 & 22.80 & 23.86 & 49.83 & 63.14 & 3.97 \\
    GradDrop & 39.39 & 65.12 & 0.5455 & 0.2279 & 27.48 & 22.96 & 23.38 & 49.44 & 62.87 & 3.58 \\
    IMTL-G & 39.35 & 65.60 & 0.5426 & 0.2256 & 26.02 & 21.19 & 26.20 & 53.13 & 66.24 & -0.76 \\
    CAGrad & 39.79 & 65.49 & 0.5486 & 0.2250 & 26.31 & 21.58 & 25.61 & 52.36 & 65.58 & 0.20 \\
    MoCo & 40.30 & \cellcolor{lightgray}\textbf{66.07} & 0.5575 & 0.2135 & 26.67 & 21.83 & 25.61 &  51.78 & 64.85 & 0.16 \\
    Nash-MTL & 40.13 & 65.93 & \cellcolor{lightgray}\textbf{0.5261} & 0.2171 & 25.26 & 20.08 & 28.40 & 55.47 & 68.15 & -4.04 \\
    FAMO & 38.88 & 64.90 & 0.5474 & 0.2194 & 25.06 & 19.57 & 29.21 & 56.61 & 68.98 & -4.10 \\
    FairGrad & 38.80 & 65.29 & 0.5572 & 0.2322 & 24.55 & 18.97 & 30.50 & 57.94 & 70.14 & -4.96 \\
    F-MTL$^\sharp$ & \cellcolor{lightgray}\textbf{40.42} & 65.61 & 0.5389 & \cellcolor{lightgray}\textbf{0.2121} & 25.03 & 19.75 & 28.90 & 56.19 & 68.72 & -4.77 \\
    \midrule
    SAMO-LS & 39.59 & 65.72 & 0.5514 & 0.2246 & 27.38 & 22.78 & 24.09 & 49.82 & 63.01 & 2.88 \\
    SAMO-MGDA & 29.85 & 60.83 & 0.6111 & 0.2388 & \cellcolor{lightgray}\textbf{24.11} & \cellcolor{lightgray}\textbf{18.18} & \cellcolor{lightgray}\textbf{32.16} & \cellcolor{lightgray}\textbf{59.59} & \cellcolor{lightgray}\textbf{71.15} & -2.19\\
    SAMO-FairGrad & 39.05 & 65.06 & 0.5359 & 0.2137 & 24.43 & 18.79 & 30.98 & 58.35 & 70.42 & \cellcolor{lightgray}\textbf{-6.55}$^*$ \\
    \bottomrule
  \end{tabular}
  \end{adjustbox}
  \caption{Results on NYU-v2 (3-task) dataset. The best results are highlighted in \textbf{bold} with gray background. $^*$ indicates best $\Delta m\%$ result. $^\sharp$ denotes the best results obtained by F-MTL.}
  \label{tab:nyuv2}
\end{table*}

\vspace{0.1cm}
\noindent\textbf{Dense Prediction.} Cityscapes \cite{cordts2016cityscapes} is an urban street scene dataset comprising 5,000 pixel-level annotated images, supporting two tasks: 7-class semantic segmentation and depth estimation. NYU-v2 \cite{silberman2012indoor} is designed for indoor scene understanding and contains 1,449 densely annotated images, supporting three tasks: 13-class semantic segmentation, depth estimation and surface normal prediction. Following \cite{liu2021conflict,navon2022multi,liu2023famo,ban2024fair}, we employ MTAN \cite{liu2019end} as the shared backbone, with task-specific attention modules built on top of SegNet \cite{badrinarayanan2017segnet}. The model is trained for 200 epochs with a batch size of 8 for Cityscapes and 2 for NYU-v2. The learning rate is set to 1e-4 for the first 100 epochs and is then halved for the remainder.

\vspace{0.1cm}
\noindent\textbf{Classification.} CelebFaces Attributes (CelebA) \cite{liu2015deep} is a widely used dataset containing over 200K annotated images of celebrity faces, each labeled with 40 binary attributes such as smiling and wearing glasses, etc. It can be framed as a facial attribute classification problem with 40 tasks. We follow the experiment setup in \cite{liu2023famo,ban2024fair}, and employ a 9-layer CNN as the backbone and a linear head for each task. The model is trained for 15 epochs with a batch size of 256 using the Adam optimizer, with a learning rate of 3e-4. 

\vspace{0.1cm}
\noindent\textbf{Regression.} QM9 \cite{ramakrishnan2014quantum} is a widely used dataset in computational chemistry, designed for molecular property prediction. It contains over 130k organic molecules, represented as graphs with node and edge features, and includes 11 property prediction tasks. Although it is not an image or video dataset, we conduct this experiment to evaluate the performance of our method comprehensively. Following \cite{navon2022multi,liu2023famo,ban2024fair}, we use the example provided in PyTorch Geometric \cite{fey2019fast}, and split the dataset into 110k molecules for training, 10k for validation and the remaining 10k for testing. The model is trained for 300 epochs with a batch size of 120. The initial learning rate is set to 1e-3, and adjusted using a scheduler that reduces the learning rate if validation performance does not improve for 5 consecutive epochs. 

\begin{table}
\small
  \centering
  \begin{adjustbox}{width=0.80\linewidth}
  \begin{tabular}{lcc}
    \toprule
    \multirow{2}*{Method} & \multicolumn{1}{c}{CelebA (40 tasks)} & \multicolumn{1}{c}{QM9 (11 tasks)} \\ 
    \cmidrule(lr){2-2}\cmidrule(lr){3-3}
    & $\Delta m\%\downarrow$ & $\Delta m\%\downarrow$ \\
    \midrule
    LS & 4.15 & 177.6 \\
    SI & 7.20 & 77.8 \\
    RLW & 1.46 & 203.8 \\
    DWA & 3.20 & 175.3 \\
    UW & 3.23 & 108.0 \\
    MGDA & 14.85 &  120.5 \\
    PCGrad & 3.17 & 125.7 \\
    IMTL-G & 0.84 & 77.2 \\
    CAGrad & 2.48 & 112.8 \\
    Nash-MTL & 2.84 & 62.0 \\
    FAMO & 1.21 & 58.5 \\
    FairGrad & 0.37 & 57.9 \\
    \midrule
    SAMO-LS & 0.66 & 141.8 \\
    SAMO-MGDA & 9.59 & 96.8 \\
    SAMO-FairGrad & \cellcolor{lightgray}\textbf{-0.74}$^*$ & \cellcolor{lightgray}\textbf{53.0}$^*$ \\
    \bottomrule
  \end{tabular}
  \end{adjustbox}
  \caption{Results on CelebA (40-task)and QM9 (11-task) datasets. The best results are highlighted in \textbf{bold} with gray background.
  }
  \label{tab:celeba_qm9}
  \vspace{-5mm}
\end{table}

\vspace{0.1cm}
\noindent\textbf{Evaluation.} We apply the proposed approach to  LS, MGDA \cite{sener2018multi}  and FairGrad \cite{ban2024fair}. These methods are chosen because they represent typical task-balancing methods. LS naively sums all task losses, where the task with the largest gradient magnitude may dominate the optimization process. MGDA, on the other hand, seeks a common update direction that decreases all task losses by computing a convex combination of task gradients, where the task with the smallest gradient magnitude will be prioritized. Many existing methods, such as CAGrad \cite{liu2021conflict} and SDMGrad \cite{xiao2023direction}, are variants of MGDA. FairGrad extends Nash-MTL \cite{navon2022multi} from a fairness perspective and aims to determine an update direction using a conic combination of task gradients, resulting in a more balanced solution.

For classification and regression, we compare the performance against a wide range of task balancing methods, including LS, Scale-Invariant (SI) which minimizes the sum of logarithmic losses, Random Loss Weighting (RLW) \cite{linreasonable}, Dynamic Weight Average (DWA) \cite{liu2019end}, Uncertainty Weighting (UW) \cite{kendall2018multi}, MGDA \cite{sener2018multi}, PCGrad \cite{yu2020gradient}, IMTL-G \cite{liutowards}, CAGrad \cite{liu2021conflict}, Nash-MTL \cite{navon2022multi}, FAMO \cite{liu2023famo}, and FairGrad \cite{ban2024fair}. For dense prediction, we additionally compare with GradDrop \cite{chen2020just}, MoCo \cite{fernandomitigating} and F-MTL \cite{phan2022improving}. 
To evaluate the overall performance of the MTL method $m$, we adopt the metric $\Delta m\%$, which measures the average per-task performance drop relative to the single-task baseline $b$:
\begin{align*}
    \Delta m\%=\frac{1}{K}\sum_{i=1}^K (-1)^{\delta_k}\frac{M_{m,k}-M_{b,k}}{M_{b,k}} \times 100,
\end{align*}
where $M_{b,k}$ denotes the value of metric $M_k$ from baseline $b$, and $M_{m,k}$ is the corresponding value of method $m$. The indicator $\delta_k$ is set to 1 if a higher value of metric $M_k$ means better performance, and 0 otherwise.

\vspace{0.1cm}
\noindent\textbf{Results.} The results are presented in \Cref{tab:cityscapes,tab:nyuv2,tab:celeba_qm9}, respectively. For Cityscapes, SAMO improves FairGrad from the $\Delta m\%$ of 3.90 to -0.62, achieving state-of-the-art performance. Furthermore, SAMO enhances all three MTL methods across both tasks. As discussed earlier, LS allows the task with the largest gradient magnitude to dominate the optimization process, while MGDA prioritizes the task with the smallest gradient magnitude. LS outperforms MGDA in the semantic segmentation task, while MGDA surpasses LS in the depth estimation task, indicating the conflict between the two tasks. SAMO effectively improves these methods in both tasks, achieving a more balanced performance. For NYU-v2, similar findings can also be observed. Notably, SAMO-FairGrad achieves state-of-the-art performance with the $\Delta m\%$ of -6.55. Additionally, SAMO consistently enhances MTL methods across nearly all tasks. 

We report the best results of F-MTL on Cityscapes and NYU-v2, obtained by F-MGDA and F-IMTL, respectively. {\em Since F-MTL does not report results for FairGrad, we implemented F-MTL with this method.} We tested various hyperparameters for F-FairGrad, but failed to obtain comparable performance, so we do not report its results. 

For CelebA and QM9, SAMO still significantly improves all three MTL methods. SAMO-FairGrad improves $\Delta m\%$ from 0.37 to -0.74 on CelebA and from 57.9 to 53.0 on QM9. F-MTL is not included as it uses a different backbone model for CelebA and does not run on QM9.

\subsection{Multi-input MTL}
We further evaluate the proposed method on Office-Home dataset \cite{venkateswara2017deep}, which consists of 4 classification tasks across different domains: artistic images (including paintings and sketches), clipart images, product images, and real-world images captured with a camera. The dataset contains 15,500 labeled images, with each domain comprising 65 classes. 

Following the codebase LibMTL \cite{lin2023libmtl}, we randomly split the dataset into 60\% for training, 20\% for validation, and the rest 20\% for testing. We use a ResNet-18 network \cite{he2016deep} pretrained on ImageNet dataset \cite{deng2009imagenet}, concatenated with a fully connected layer,  as the shared backbone. A linear classification head is then applied for each task. 

Similar to the single-input scenario, we apply our method to LS, MGDA, and FairGrad, and compare the top-1 accuracy and the average accuracy across these tasks. The model is trained for 50 epochs with a batch size of 64. The learning rate is set to 1e-4 for the first 25 epochs and is then halved for the remaining 25 epochs.

\vspace{0.1cm}
\noindent\textbf{Results.} It can be noticed that SAMO also can improve all three MTL methods under the multi-input setting. Additionally, we observe that LS outperforms MGDA and FairGrad. We infer that, due to the use of a pretrained model, the changes in shared parameters between the original pretrained model and finetuned models are minimal. As a result, the differences among the models adapted under the three MTL methods are also small, suggesting that the cause of task conflicts is not gradient conflicts. This limitation further constrains the performance of methods like MGDA and FairGrad. However, SAMO remains effective in mitigating task conflicts even in situations where typical gradient manipulation methods fail, ultimately improving the overall performance.

\begin{table}
  \centering
  \begin{adjustbox}{width=0.95\linewidth}
  \begin{tabular}{lccccl}
    \toprule
    Method & Art & Clipart & Product & Real World & Avg Acc $\uparrow$ \\
    \midrule
    STL & 67.74 & 80.39 & 90.57 & 81.08 & 79.95 \\
    \midrule
    LS & 62.02 & 77.03 & 89.72 & \cellcolor{lightgray}\textbf{80.32} & 77.27 \\
    MGDA & 62.24 & 70.42 & \cellcolor{lightgray}\textbf{90.25} & 78.05 & 75.24 \\
    FairGrad & 62.68 & 76.49 & 89.62 & 79.68 & 77.12 \\
    \midrule
    SAMO-LS &\cellcolor{lightgray} \textbf{63.23} & \cellcolor{lightgray}\textbf{77.46} & 90.15 & 79.46 & 77.58 \\
    SAMO-MGDA & 62.13 & 71.29 & 89.94 & 78.05 & 75.35 \\
    SAMO-FairGrad & 63.12 & 77.03 & \cellcolor{lightgray}\textbf{90.25} & 80.22 & \cellcolor{lightgray}\textbf{77.66}$^*$ \\
    \bottomrule
  \end{tabular}
  \end{adjustbox}
  \vspace{-0.1cm}
  \caption{Results on Office Home (4-task) dataset. The best results are highlighted in \textbf{bold} with gray background.}
  \label{tab:office_home}
  \vspace{-0.3cm}
\end{table}

\subsection{Ablation study}
We compare our SAMO with SAM that uses only global or local information, and present the results of MGDA and FairGrad, on Cityscapes, NYU-v2, and CelebA datasets in \Cref{tab:ablation}. The results indicate that global and local information can enhance the MTL methods. However, SAMO utilizes a joint perturbation and generally achieves better performance. We also conduct an ablation study on perturbation normalization, which supports our choice of layerwise normalization. For more details, please refer to \Cref{app:ablation}.

\begin{table}
  \centering
  \small
  \begin{adjustbox}{width=0.85\linewidth}
  \begin{tabular}{lccc}
    \toprule
    \multirow{2}*{Method} & \multicolumn{1}{c}{Cityscapes} & \multicolumn{1}{c}{NYU-v2} & \multicolumn{1}{c}{CelebA} \\ 
     
    \cmidrule(lr){2-2}\cmidrule(lr){3-3}\cmidrule(lr){4-4}
    & $\Delta m\%\downarrow$ & $\Delta m\%\downarrow$ & $\Delta m\%\downarrow$ \\
    \midrule
    MGDA & 44.14 & 1.38 & 14.85 \\
    G-SAM-MGDA & 7.51 & -0.23 & 11.78 \\
    L-SAM-MGDA & 11.94 & 0.01 & \cellcolor{lightgray}\textbf{8.47} \\
    SAMO-MGDA & \cellcolor{lightgray}\textbf{4.30} & \cellcolor{lightgray}\textbf{-2.19} & 9.59 \\
    \midrule
    FairGrad & 3.90 & -4.96 & 0.37  \\
    G-SAM-FairGrad & 0.93 & -5.70 & 0.41 \\
    L-SAM-FairGrad & 1.01 & -5.42 & -0.42  \\
    SAMO-FairGrad & \cellcolor{lightgray}\textbf{-0.62} & \cellcolor{lightgray}\textbf{-6.55} & \cellcolor{lightgray}\textbf{-0.74} \\
    \bottomrule
  \end{tabular}
  \end{adjustbox}
  \vspace{-0.1cm}
  \caption{Ablation study of SAMO on Cityscapes, NYU-v2, and CelebA datasets. The best results of each method are highlighted in \textbf{bold} with gray background.}
  \label{tab:ablation}
  \vspace{-0.4cm}
\end{table}

\subsection{Efficiency comparison}
\begin{figure}[ht]
\vspace{-0.2cm}
  \centering
   \includegraphics[width=\linewidth]{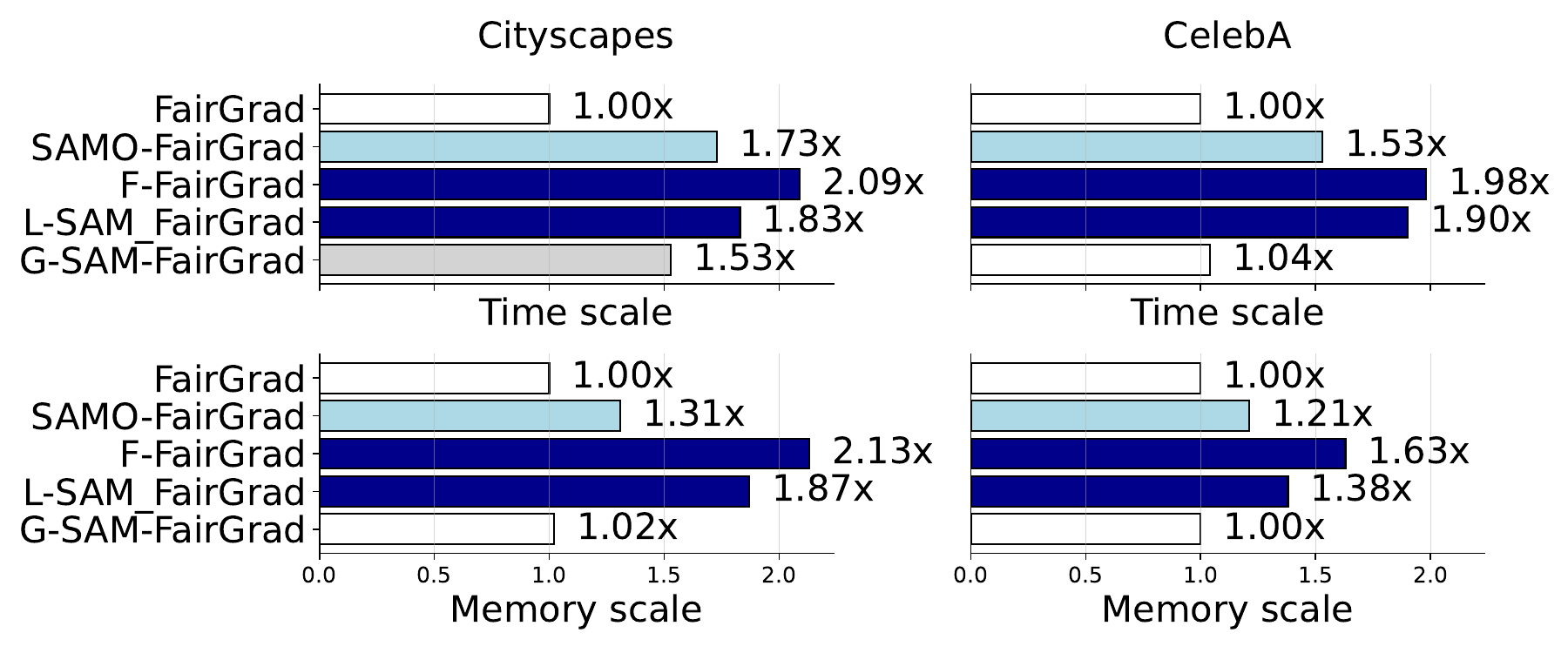}
      \vspace{-0.6cm}
   \caption{Efficiency comparison on Cityscapes (2-task) and CelebA (40-task) datasets.}
   \label{fig:efficiency}
   \vspace{-0.4cm}
\end{figure}

We also compare our method with global and local SAM, and F-MTL, in terms of running time and memory cost. Since SAM introduces operations that perturbate model parameters, which have different implementations and obscure the running time comparisons. Therefore, we primarily focus on computing time in the running time comparison. We conducted all the experiments on an RTX A6000 GPU. The results are shown in \Cref{fig:efficiency}. Our SAMO exhibits significant improvements in both computing time and memory cost compared to L-SAM and F-MTL that {\bf utilize local information}.

\section{Conclusion}

We first present empirical findings showing that SAM mitigates task conflicts in MTL, with both average loss and individual task gradients playing key roles. We then examine practical strategies for integrating SAM, noting limitations in existing methods. To address these, we propose SAMO, a lightweight approach that balances global and local information efficiently. Finally, we validate SAMO’s effectiveness through extensive experiments. For future studies, we will explore more scalable and efficient versions of SAMO through sparsity or low-rank approximation.

\section*{Acknowledgement}

We sincerely thank all the reviewers for providing valuable and constructive feedback. This work is generously supported by the NSF Career Award under Grant No. 2442418.
{
    \small
    \bibliographystyle{ieeenat_fullname}
    \bibliography{main}
}

\clearpage
\setcounter{page}{1}
\maketitlesupplementary

\appendix

\section{Sharpness and task conflicts}\label{app:sharpness}
For the experiment in \Cref{sec:sam_conflict}, we use the same model and experimental setup as in \Cref{sec:experiments}. Following \citealt{foretsharpness}, we remove batch normalization, as it affects the interpretation of Hessian. We provide the cosine similarities of task gradients for LS and LS with SAM after the 5th and 10th epoch, corresponding to one-third and two-thirds of the training progression, respectively. 



\begin{figure}[H]
    \centering
    \includegraphics[width=0.45\linewidth]{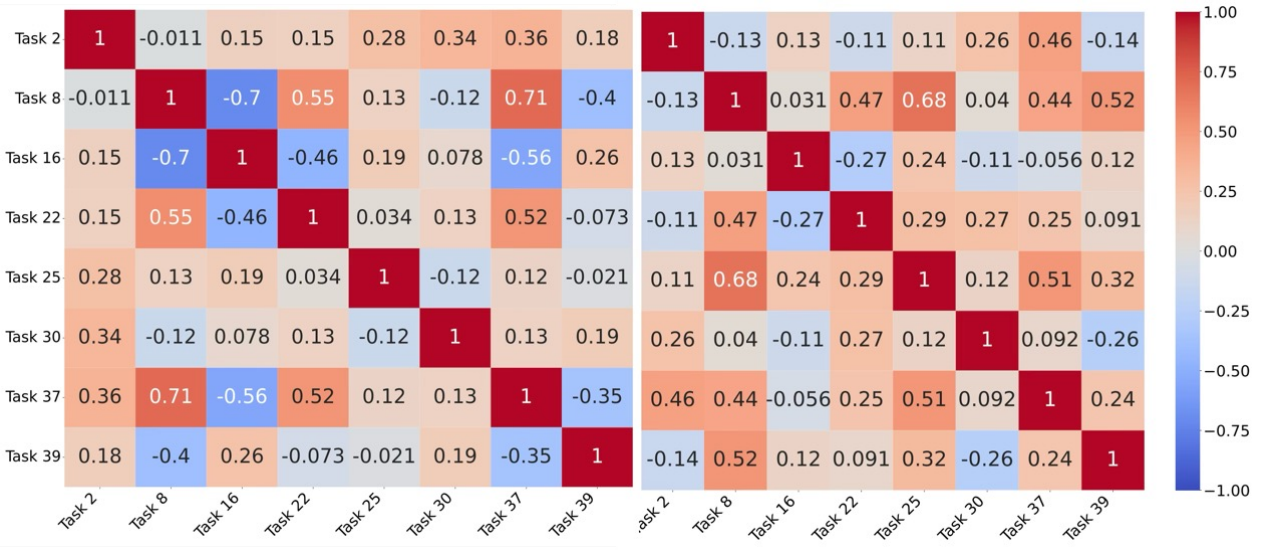}
    \hspace{-1.5mm}
    \includegraphics[width=0.45\linewidth]{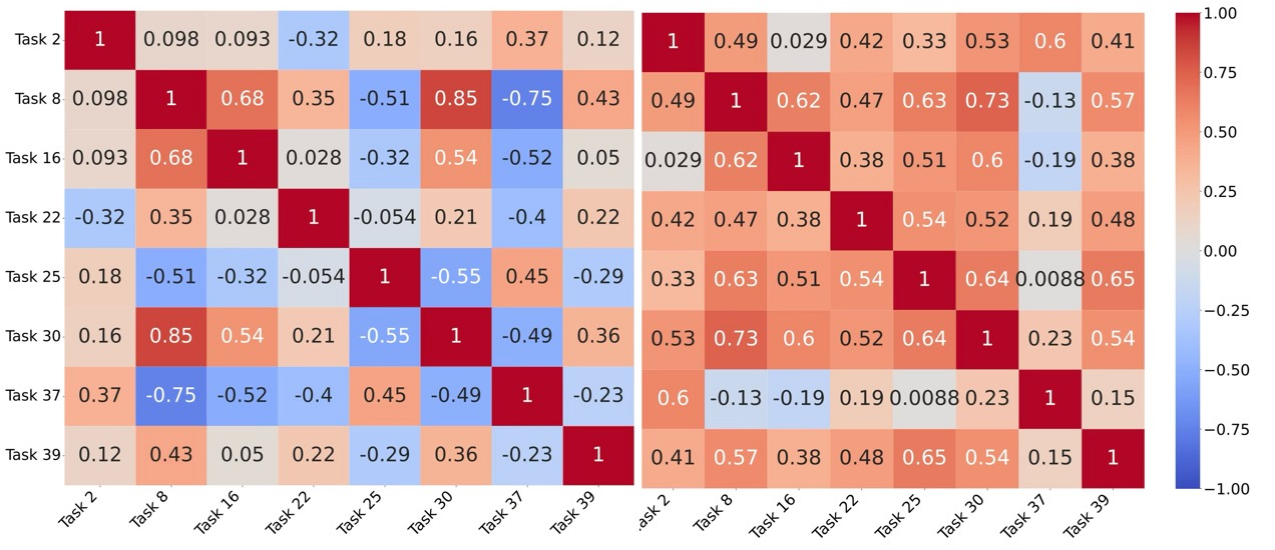}
    \caption{Cosine similarities of task gradients after 5th (left) and 10th (right) epoch. In each figure, LS is on the left and LS with SAM is on the right.}
    \label{fig:app_cosine}
\end{figure}

\section{Hyperparameter selection and sensitivity analysis}\label{app:hyperparameter}
For each dataset, we first search $\rho$ within a wide range, $[0.00001, 0.0001, 0.001, 0.01, 0.1]$, and evaluate the best choice. We then refine the search space to determine an optimal value. For Cityscapes, NYU-v2, CelebA, Office Home, we use $\rho=0.001$ and $\alpha=0.5$. For QM9, the loss scales vary across different tasks; therefore, the parameter $\rho$ should be carefully selected. We set $\rho=0.00001$, and $\alpha=0.1$. We set $\mu=0.01$ in \Cref{eq:zeroth_order} to approximate each task gradient using only forward passes. 

Furthermore, we conduct the sensitivity analysis of $\alpha$ on Cityscapes, NYU-v2, QM9, and Office Home datasets. The results below indicate that model's performance is not highly sensitive to $\alpha$. Empirically, we observe that the model remains relatively robust within the range $\alpha\in[0.1,0.6]$.

In Addition, we also perform a sensitivity analysis of $\rho$ on Cityscapes and NYU-v2 datasets. We explore values $\rho\in \{0.0001, 0.01\}$. We observe from \Cref{tab:sensitivity_rho} that model performance is more sensitive to $\rho$ mainly because it plays a crucial role in the behavior of SAM.

\begin{table}[H]
\centering
\begin{adjustbox}{width=0.6\linewidth}
\begin{tabular}{lcccc}
\toprule
Method & Cityscapes $\downarrow$ & NYU-v2 $\downarrow$ & QM9 $\downarrow$  & Office $\uparrow$ \\
\midrule
LS & 22.60 & 5.59 & 177.6 & 77.27 \\
SAMO-LS ($\alpha=0.1$) & 18.84 & 3.28 & \cellcolor{lightgray}141.8 & 77.14 \\
SAMO-LS ($\alpha=0.5$) & \cellcolor{lightgray}14.30 & \cellcolor{lightgray}2.88 & 160.0 & \cellcolor{lightgray}77.58 \\
SAMO-LS ($\alpha=0.9$) & 24.81 & 6.38 & 213.9 & 76.68 \\
\midrule
FairGrad & 3.90 & -4.96 & 57.9 & 77.12 \\
SAMO-FairGrad ($\alpha=0.1$) & 1.45 & -5.28 & \cellcolor{lightgray}53.0 & 76.65 \\
SAMO-FairGrad ($\alpha=0.5$) & \cellcolor{lightgray}-0.62 & \cellcolor{lightgray}-6.55 & 55.7 & \cellcolor{lightgray}77.66 \\
SAMO-FairGrad ($\alpha=0.9$) & 6.49 & -4.56 & 68.5 & 76.29\\
\bottomrule
\end{tabular}
\end{adjustbox}
\caption{Sensitivity of $\alpha$. The reported results are highlighted.}
\label{tab:sensitivity_alpha}
\end{table}

\begin{table}[H]
\centering
\begin{adjustbox}{width=0.5\linewidth}
\begin{tabular}{lcc}
\toprule
Method & Cityscapes $\downarrow$ & NYU-v2 $\downarrow$ \\
\midrule
LS & 22.60 & 5.59 \\
SAMO-LS ($\rho=0.01$) & 24.03 & 3.59 \\
\cellcolor{lightgray}SAMO-LS ($\rho=0.001$) & \cellcolor{lightgray}14.30 & \cellcolor{lightgray}2.88 \\
SAMO-LS ($\rho=0.0001$) & 18.78 & 3.67 \\
\midrule
FairGrad & 3.90 & -4.96 \\
SAMO-FairGrad ($\rho=0.01$) & 14.03 & -4.38 \\
\cellcolor{lightgray}SAMO-FairGrad ($\rho=0.001$) & \cellcolor{lightgray}-0.62 & \cellcolor{lightgray}-6.55 \\
SAMO-FairGrad ($\rho=0.0001$) & 6.11 & -4.15 \\
\bottomrule
\end{tabular}
\end{adjustbox}
  \caption{Sensitivity of $\rho$. The reported results are highlighted.}
\label{tab:sensitivity_rho}
\end{table}

\section{Detailed results on multi-task regression}\label{app:full_qm9}
We provide more details about per-task results on the QM9 dataset in \Cref{tab:full_qm9}.

\begin{table}[H]
\centering
\begin{adjustbox}{width=0.8\linewidth}
  \begin{tabular}{lccccccccccccc}
    \toprule
    \multirow{2}*{Method} & $\mu$ & $\alpha$ & $\epsilon_{HOMO}$ & $\epsilon_{LUMO}$ & $\langle R^2\rangle$ & ZPVE & $U_0$ & $U$ & $H$ & $G$ & $c_v$ & \multirow{2}*{$\Delta m\%\downarrow$} \\
    \cmidrule(lr){2-12}
    & \multicolumn{11}{c}{MAE $\downarrow$} & & \\
    \midrule
    STL & 0.067 & 0.181 & 60.57 & 53.91 & 0.502 & 4.53 & 58.8 & 64.2 & 63.8 & 66.2 & 0.072 & & \\
    \midrule
    LS & 0.106 & 0.325 & 73.57 & 89.67 & 5.19 & 14.06 & 143.4 & 144.2 & 144.6 & 140.3  & 0.128 & 177.6 \\
    SI & 0.309 & 0.345 & 149.8 & 135.7 & \cellcolor{lightgray}\textbf{1.00} & 4.50 & \cellcolor{lightgray}\textbf{55.3} & \cellcolor{lightgray}\textbf{55.75} & \cellcolor{lightgray}\textbf{55.82} & \cellcolor{lightgray}\textbf{55.27}  & 0.112 & 77.8 \\
    RLW & 0.113 & 0.340 & 76.95 & 92.76 & 5.86 & 15.46 & 156.3 & 157.1 & 157.6 & 153.0  & 0.137 & 203.8 \\
    DWA & 0.107 & 0.325 & 74.06 & 90.61 & 5.09 & 13.99 & 142.3 & 143.0 & 143.4 & 139.3  & 0.125 & 175.3 \\
    UW & 0.386 & 0.425 & 166.2 & 155.8 & 1.06 & 4.99 & 66.4 & 66.78 & 66.80 & 66.24  & 0.122 & 108.0 \\
    MGDA & 0.217 & 0.368 & 126.8 & 104.6 & 3.22 & 5.69 & 88.37 & 89.4 & 89.32 & 88.01  & 0.120 & 120.5 \\
    PCGrad & 0.106 & 0.293 & 75.85 & 88.33 & 3.94 & 9.15 & 116.36 & 116.8 & 117.2 & 114.5  & 0.110 & 125.7 \\
    IMTL-G & 0.136 & 0.287 & 98.31 & 93.96 & 1.75 & 5.69 & 101.4 & 102.4 & 102.0 & 100.1  & 0.096 & 77.2 \\
    CAGrad & 0.118 & 0.321 & 83.51 & 94.81 & 3.21 & 6.93 & 113.99 & 114.3 & 114.5 & 112.3  & 0.116 & 112.8 \\
    Nash-MTL & 0.102 & \cellcolor{lightgray}\textbf{0.248} & 82.95 & \cellcolor{lightgray}\textbf{81.89} & 2.42 & 5.38 & 74.50 & 75.02 & 75.10 & 74.16  & \cellcolor{lightgray}\textbf{0.093} & 62.0 \\
    FAMO & 0.150 & 0.300 & 94.00 & 95.20 & 1.63 & 4.95 & 70.82 & 71.20 & 71.20 & 70.30 & 0.100 & 58.5 \\
    FairGrad & 0.117 & 0.253 & 87.57 & 84.00 & 2.15 & 5.07 & 70.89 & 71.17 & 71.21 & 70.88 & 0.095 & 57.9 \\
    \midrule
    SAMO-LS & \cellcolor{lightgray}\textbf{0.096} & 0.301 & \cellcolor{lightgray}\textbf{66.66} & 82.81 & 4.69 & 11.16 & 117.49 & 118.11 & 118.56 & 116.86 & 0.117 & 141.8 \\
    SAMO-MGDA & 0.203 & 0.323 & 119.58 & 86.63 & 2.21 & 5.35 & 97.80 & 98.53 & 98.26 & 98.10 & 0.107 & 96.8 \\
    SAMO-FairGrad & 0.137 & 0.255 & 99.53 & 87.31 & 1.72 & \cellcolor{lightgray}\textbf{4.30} & 70.39 & 70.88 & 70.70 & 70.98 & \cellcolor{lightgray}\textbf{0.093} & \cellcolor{lightgray}\textbf{53.0} \\
    \bottomrule
  \end{tabular}
  \end{adjustbox}
  \caption{Detailed results of  on QM9 (11-task) dataset. The best results are highlighted in \textbf{bold} with gray background.}
\label{tab:full_qm9}
\end{table}

\section{Ablation study}\label{app:ablation}
We provide more detailed per-task results of ablation study on Cityscapes, and NYU-v2 datasets. The results are presented in \Cref{tab:ablation_cityscapes} and \Cref{tab:ablation_nyuv2}.

\begin{table}[H]
  \centering
  \begin{adjustbox}{width=0.5\linewidth}
  \begin{tabular}{lccccc}
    \toprule
    \multirow{2}*{Method} & \multicolumn{2}{c}{Segmentation} & \multicolumn{2}{c}{Depth} & 
    \multirow{2}*{$\Delta m\%\downarrow$} \\
    \cmidrule(lr){2-3}\cmidrule(lr){4-5}
    & mIoU $\uparrow$ & Pix Acc $\uparrow$ & Abs Err $\downarrow$ & Rel Err $\downarrow$ & \\
    \midrule
    LS & 75.18 & 93.49 & 0.0155 & 46.77 & 22.60 \\
    G-SAM-LS & 75.78 & 93.59 & 0.0155 & 41.68 & 17.85 \\
    L-SAM-LS & 74.48 & 93.34 & 0.0154 & 40.07 & 16.71 \\
    SAMO-LS & \cellcolor{lightgray}\textbf{76.46} & \cellcolor{lightgray}\textbf{93.76} & \cellcolor{lightgray}\textbf{0.0147} & \cellcolor{lightgray}\textbf{39.85} & \cellcolor{lightgray}\textbf{14.30} \\
    \midrule
    MGDA & 68.84 & 91.54 & 0.0309 & 33.50 & 44.14 \\
    G-SAM-MGDA & 72.64 & 92.97 & 0.0150 & 30.01 & 7.51 \\
    L-SAM-MGDA & 72.77 & 92.50 & 0.0185 & \cellcolor{lightgray}\textbf{26.97} & 11.94 \\
    SAMO-MGDA & \cellcolor{lightgray}\textbf{73.28} & \cellcolor{lightgray}\textbf{93.26} & \cellcolor{lightgray}\textbf{0.0133} & 30.57 & \cellcolor{lightgray}\textbf{4.30} \\
    \midrule
    FairGrad & 74.10 & 93.03 & 0.0135 & 29.92 & 3.90 \\
    G-SAM-FairGrad & \cellcolor{lightgray}\textbf{74.81} & 93.12 & \cellcolor{lightgray}\textbf{0.0126} & 28.85 & 0.93 \\
    L-SAM-FairGrad & 74.16 & 93.11 & 0.0135 & 26.60 & 1.01 \\
    SAMO-FairGrad & 74.37 & \cellcolor{lightgray}\textbf{93.14} & 0.0129 & \cellcolor{lightgray}\textbf{26.30} & \cellcolor{lightgray}\textbf{-0.62} \\
    \bottomrule
  \end{tabular}
  \end{adjustbox}
  \caption{Ablation study of SAMO on Cityscapes (2-task) dataset. The best results of each method are highlighted in \textbf{bold} with gray background.}
  \label{tab:ablation_cityscapes}
\end{table}

\begin{table}[H]
  \centering
  \begin{adjustbox}{width=0.75\linewidth}
  \begin{tabular}{llllllllllc}
    \toprule
    \multirow{3}*{Method} & \multicolumn{2}{c}{Segmentation} & \multicolumn{2}{c}{Depth} & \multicolumn{5}{c}{Surface Normal} & \multirow{3}*{$\Delta m\%\downarrow$} \\
    \cmidrule(lr){2-3}\cmidrule(lr){4-5}\cmidrule(lr){6-10}
    & \multirow{2}*{mIoU $\uparrow$} & \multirow{2}*{Pix Acc $\uparrow$} & \multirow{2}*{Abs Err $\downarrow$} & \multirow{2}*{Rel Err $\downarrow$} & \multicolumn{2}{c}{Angle Distance $\downarrow$} & \multicolumn{3}{c}{Within $t^\circ$ $\uparrow$} & \\
    \cmidrule(lr){6-7}\cmidrule(lr){8-10}
    & & & & & Mean & Median & $<$11.25 & $<$22.5 & $<$30 & \\
    \midrule
    LS & 39.29 & 65.33 & 0.5493 & 0.2263 & 28.15 & 23.96 & 22.09 & 47.50 & 61.08 & 5.59 \\
    G-SAM-LS & 39.17 & 64.84 & 0.5500 & 0.2301 & 27.47 & 23.02 & 23.60 & 49.33 & 62.74 & 3.85 \\
    L-SAM-LS & 38.14 & 64.76 & \cellcolor{lightgray}\textbf{0.5324} & \cellcolor{lightgray}\textbf{0.2215} & \cellcolor{lightgray}\textbf{27.12} & \cellcolor{lightgray}\textbf{22.40} & \cellcolor{lightgray}\textbf{25.04} & \cellcolor{lightgray}\textbf{50.50} & \cellcolor{lightgray}\textbf{63.59} & \cellcolor{lightgray}\textbf{2.11} \\
    SAMO-LS & \cellcolor{lightgray}\textbf{39.59} & \cellcolor{lightgray}\textbf{65.72} & 0.5514 & 0.2246 & 27.38 & 22.78 & 24.09 & 49.82 & 63.01 & 2.88 \\
    \midrule
    MGDA & \cellcolor{lightgray}\textbf{30.47} & 59.90 & \cellcolor{lightgray}\textbf{0.6070} & 0.2555 & 24.88 & 19.45 & 29.18 & 56.88 & 69.36 & 1.38 \\
    G-SAM-MGDA & 29.49 & 59.60 & 0.6107 & \cellcolor{lightgray}\textbf{0.2349} & 24.77 & 18.94 & 30.77 & 57.90 & 69.75 & -0.23 \\
    L-SAM-MGDA & 27.44 & 59.89 & 0.6370 & 0.2356 & 24.53 & 18.61 & 31.42 & 58.59 & 70.33 & 0.01 \\
    SAMO-MGDA & 29.85 & \cellcolor{lightgray}\textbf{60.83} & 0.6111 & 0.2388 & \cellcolor{lightgray}\textbf{24.11} & \cellcolor{lightgray}\textbf{18.18} & \cellcolor{lightgray}\textbf{32.16} & \cellcolor{lightgray}\textbf{59.59} & \cellcolor{lightgray}\textbf{71.15} & \cellcolor{lightgray}\textbf{-2.19}\\
    \midrule
    FairGrad & 38.80 & 65.29 & 0.5572 & 0.2322 & 24.55 & 18.97 & 30.50 & 57.94 & 70.14 & -4.96 \\
    G-SAM-FairGrad & 37.90 & 65.08 & \cellcolor{lightgray}\textbf{0.5269} & 0.2162 & 24.50 & 19.03 & 30.38 & 57.76 & 70.04 & -5.70 \\
    L-SAM-FairGrad & 38.72 & \cellcolor{lightgray}\textbf{65.42} & 0.5737 & 0.2455 & \cellcolor{lightgray}\textbf{24.16} & \cellcolor{lightgray}\textbf{18.43} & \cellcolor{lightgray}\textbf{31.67} & \cellcolor{lightgray}\textbf{59.06} & \cellcolor{lightgray}\textbf{70.92} & -5.42 \\
    SAMO-FairGrad & \cellcolor{lightgray}\textbf{39.05} & 65.06 & 0.5359 & \cellcolor{lightgray}\textbf{0.2137} & 24.43 & 18.79 & 30.98 & 58.35 & 70.42 & \cellcolor{lightgray}\textbf{-6.55} \\
    \bottomrule
  \end{tabular}
  \end{adjustbox}
  \vspace{-2mm}
  \caption{Ablation study of SAMO on NYU-v2 (3-task) dataset. The best results of each method are highlighted in \textbf{bold} with gray background.}
  \label{tab:ablation_nyuv2}
\vspace{-2mm}
\end{table}

Moreover, we conduct ablation study on the perturbation normalization and present the results in \Cref{tab:ablation}. Global normalization scales the local perturbation for task $i$ by the factor $\frac{\|\nabla_\theta l_0(\theta)\|}{\|\hat{\nabla}_\theta l_i(\theta)\|}$, where numerator and denominator are the norms of global and local perturbations across all layers, respectively. It can be observed that our layer-wise normalization consistently outperforms alternative methods.

\vspace{-1mm}

\begin{table}[H]
  \centering
  \begin{adjustbox}{width=0.70\linewidth}
  \begin{tabular}{lllllcc}
    \toprule
    \multirow{3}*{Method} & \multicolumn{5}{c}{Cityscapes} & \multicolumn{1}{c}{NYU-v2}\\
    \cmidrule(lr){2-6}\cmidrule(lr){7-7}
    & \multicolumn{2}{c}{Segmentation} & \multicolumn{2}{c}{Depth} & 
    \multirow{2}*{$\Delta m\%\downarrow$} & \multirow{2}*{$\Delta m\%\downarrow$} \\
    \cmidrule(lr){2-3}\cmidrule(lr){4-5}
    & mIoU $\uparrow$ & Pix Acc $\uparrow$ & Abs Err $\downarrow$ & Rel Err $\downarrow$ & \\
    \midrule
    LS & 75.18 & 93.49 & 0.0155 & 46.77 & 22.60 & 5.59 \\
    \cellcolor{lightgray}SAMO-LS & \cellcolor{lightgray}76.46 & \cellcolor{lightgray}93.76 & \cellcolor{lightgray}0.0147 & \cellcolor{lightgray}39.85 & \cellcolor{lightgray}\textbf{14.30} & \cellcolor{lightgray}\textbf{2.88} \\
    SAMO-LS (glob. norm.) & 75.79 & 93.69 & 0.0142 & 44.00 & 17.21 & 4.79 \\
    SAMO-LS (w/o norm.) & 75.82 & 93.76 & 0.0143 & 43.38 & 16.80 & 4.91 \\
    \midrule
    FairGrad & 74.10 & 93.03 & 0.0135 & 29.92 & 3.90 & -4.96 \\
    \cellcolor{lightgray}SAMO-FairGrad & \cellcolor{lightgray}74.37 & \cellcolor{lightgray}93.14 & \cellcolor{lightgray}0.0129 & \cellcolor{lightgray}26.30 & \cellcolor{lightgray}\textbf{-0.62} & \cellcolor{lightgray}\textbf{-6.55} \\
    SAMO-FairGrad (glob. norm.) & 74.54 & 93.10 & 0.0130 & 27.35 & 0.42 & -5.44 \\
    SAMO-FairGrad (w/o norm.) & 74.23 & 93.02 & 0.0127 & 26.98 & -0.27 & -4.72 \\
    \bottomrule
  \end{tabular}
  \end{adjustbox}
  \vspace{-2mm}
  \caption{Ablation study of perturbation normalization on Cityscapes and NYU-v2 datasets. glob. norm. refers to global normalization of local perturbations, while w/o norm. means no normalization.}
  \label{tab:ablation_normalization}
\vspace{-2mm}
\end{table}

\section{Feature visualization}

For NYU-v2 dataset, we extract the intermediate features from the task-specific heads and visualize them using t-SNE \cite{van2008visualizing}. SAMO-FairGrad achieves a higher inter-cluster distance (52.35 vs. 50.48), indicating improved task feature separability.

\begin{figure}[h!]
\vspace{-2mm}
  \centering
\includegraphics[width=0.4\linewidth]{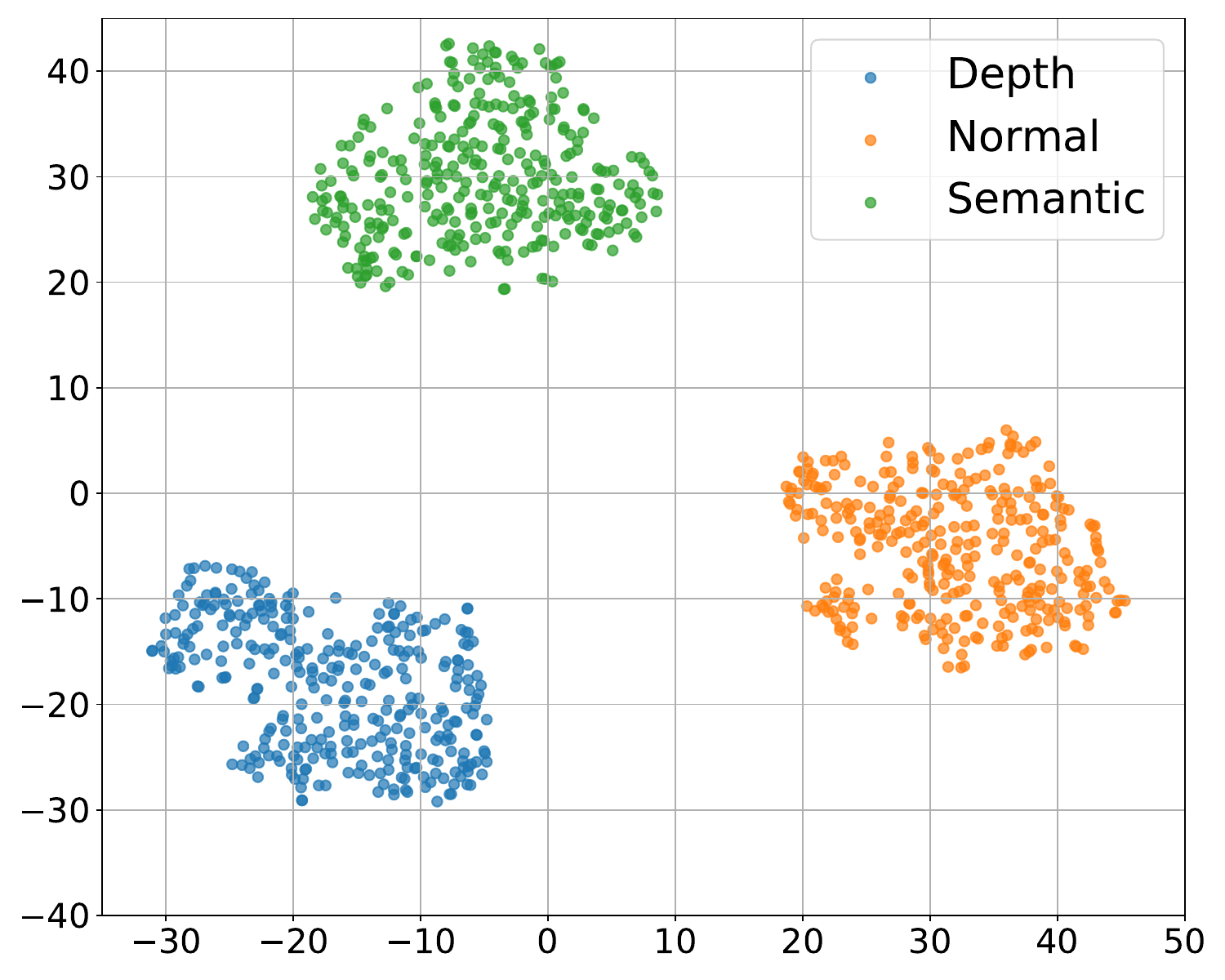}
   \hspace{-0.5mm}
   \includegraphics[width=0.4\linewidth]{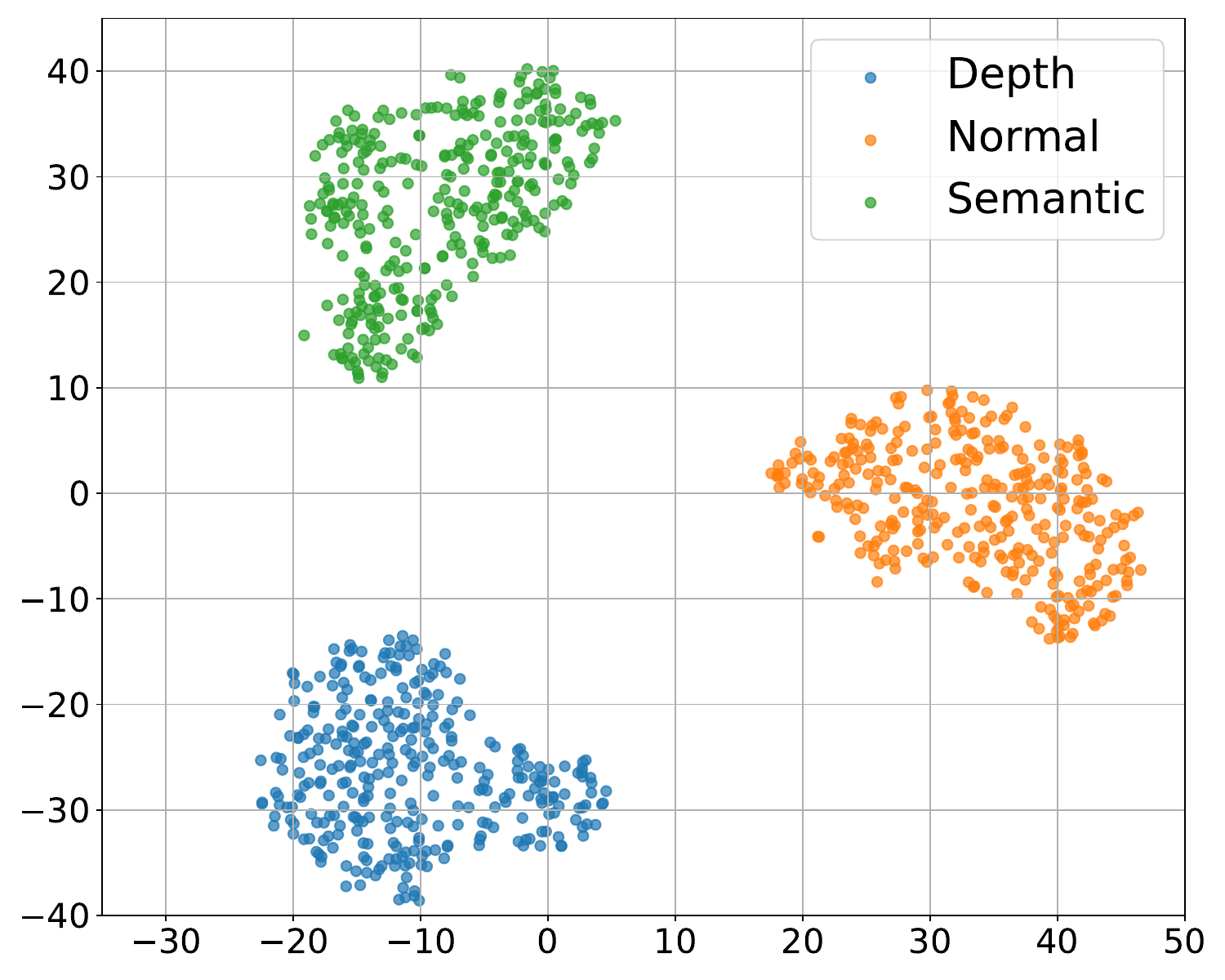}
   \vspace{-2mm}
   \caption{Feature visualization. Left: FairGrad . Right: SAMO-FairGrad.}
   \label{fig:feature}
\end{figure}

\end{document}